\documentclass[conference,11pt]{IEEEtran}

\usepackage{cite}
\usepackage{amsmath,amssymb,amsfonts}
\usepackage[caption=false,font=footnotesize]{subfig}
\usepackage{algorithmic}
\usepackage{graphicx}
\usepackage{textcomp}
\usepackage{xcolor}
\usepackage{tikz} 
\usepackage{float}
\usepackage{amsmath}
\usepackage{url}
\usepackage{hyperref}

\usetikzlibrary{arrows.meta,positioning,shapes.geometric,fit}

\def\BibTeX{{\rm B\kern-.05em{\sc i\kern-.025em b}\kern-.08em
    T\kern-.1667em\lower.7ex\hbox{E}\kern-.125emX}}

\begin{document}

\title{Interpreting Negation in GPT-2: Layer- and Head-Level Causal Analysis}

\author{
\IEEEauthorblockN{
Abdullah Al Mofael\IEEEauthorrefmark{1},
Lisa M. Kuhn\IEEEauthorrefmark{2},
Ghassan Alkadi\IEEEauthorrefmark{1},
and Kuo-Pao Yang\IEEEauthorrefmark{1}
}

\IEEEauthorblockA{\IEEEauthorrefmark{1}Department of Computer Science,
Southeastern Louisiana University, Hammond, LA, USA\\
Email:  abdullahal.mofael@southeastern.edu,
galkadi@selu.edu,
kuo-pao.yang@southeastern.edu}

\IEEEauthorblockA{\IEEEauthorrefmark{2}Department of Mathematics,
Southeastern Louisiana University, Hammond, LA, USA\\
Email: lisa.kuhn@southeastern.edu}
}

\maketitle

\newcommand\IEEENotice{%
  \begin{tikzpicture}[remember picture,overlay]
    \node[anchor=south,yshift=10pt] at (current page.south) {
      \parbox{\dimexpr\textwidth\relax}{
        \scriptsize \copyright~2026 IEEE. Personal use of this material is permitted. 
        Permission from IEEE must be obtained for all other uses, in any current or 
        future media, including reprinting/republishing this material for advertising 
        or promotional purposes, creating new collective works, for resale or 
        redistribution to servers or lists, or reuse of any copyrighted component of 
        this work in other works. This article has been accepted for publication in 
        the 2026 IEEE 16th Annual Computing and Communication Workshop and Conference 
        (CCWC). The final published version is available at: 
        \url{https://doi.org/10.1109/CCWC67433.2026.11393646}
      }
    };
  \end{tikzpicture}%
}

\IEEENotice

\begin{abstract}
Negation remains a persistent challenge for modern language models, often causing reversed meanings or factual errors. In this work, we conduct a causal analysis of how GPT-2 Small internally processes such linguistic transformations. We examine its hidden representations at both the layer and head level. Our analysis is based on a self-curated 12,000-pair dataset of matched affirmative and negated sentences, covering multiple linguistic templates and forms of negation. To quantify this behavior, we define a metric, the Negation Effect Score (NES), which measures the model’s sensitivity in distinguishing between affirmative statements and their negations. We carried out two key interventions to probe causal structure. In activation patching, internal activations from affirmative sentences were inserted into their negated counterparts to see how meaning shifted. In ablation, specific attention heads were temporarily disabled to observe how logical polarity changed. Together, these steps revealed how negation signals move and evolve through GPT-2’s layers. Our findings indicate that this capability is not widespread; instead, it is highly concentrated within a limited number of mid-layer attention heads, primarily within layers 4 to 6. Ablating these specific components directly disrupts the model’s negation sensitivity: on our in-domain, ablation increased NES (indicating weaker negation sensitivity), and re-introducing cached affirmative activations (rescue) increased NES further, confirming that these heads carry affirmative signal rather than restoring baseline behavior. On xNot360, ablation slightly decreased NES and rescue restored performance above baseline. This pattern demonstrates that these causal patterns are consistent across various negation forms (‘never,’ ‘does not,’ ‘doesn’t,’ ‘cannot,’ and ‘can’t’) and remain detectable on the external xNot360 benchmark, though with smaller magnitude. These findings offer direct evidence of a localized and interpretable circuit for logical polarity, providing a mechanistic explanation for how this transformer architecture encodes meaning.
\end{abstract}


\section{Introduction}

Negation is a subtle linguistic phenomenon that most humans process without a second thought, but for language models, it remains a consistent stumbling block. Large pretrained transformers such as GPT-2~\cite{radford2019language} and GPT-3~\cite{brown2020language} can generate impressively coherent text, yet they often falter when a single word like ``not'' reverses the meaning of a sentence~\cite{truong2023analysis}. A model might claim that ``Paris is not the capital of France'' is true or fail to distinguish between ``the cat is alive'' and ``the cat is not alive.'' These errors, although seemingly trivial, reveal deep cracks in how models encode and analyze logical structure.

Negation is not just a linguistic curiosity; it is a cornerstone of reasoning and factual reliability. In safety-critical contexts such as medical diagnosis, legal interpretation, or policy generation, a single misread negation can completely invert the truth. This has driven researchers to investigate \emph{how} transformer models represent negation internally. Behavioral benchmarks such as xNot360~\cite{nguyen2023xnot360} provide quantitative evidence that even modern models struggle to generalize negation across syntactic forms, yet those datasets alone do not explain \emph{why} these failures occur.

Recent advances in mechanistic interpretability attempt to bridge that gap. Instead of treating neural models as black boxes, interpretability research dissects their inner workings, mapping neurons, layers, and attention heads to linguistic or reasoning functions. Techniques such as \textit{activation patching} and \textit{causal ablation} have emerged as powerful diagnostic tools~\cite{heimersheim2024useinterpretactivationpatching, wang2022indirect}. By selectively modifying or substituting activations, one can observe how internal components contribute to the final output, effectively turning deep networks into traceable computational systems.

Building on this foundation, our work performs a systematic causal analysis of negation within GPT-2 Small. We use GPT-2 Small as our base model because its moderate size allows full-layer interpretability and efficient causal intervention while still exhibiting the same architectural behaviors found in larger transformer models~\cite{elhage2021mathematical, wang2022indirect}. Our goal is not to evaluate the current frontier of negation performance, but to
provide a detailed mechanistic case study on a canonical, widely used model.
Although newer large language models (e.g., GPT-3–style systems) achieve higher
behavioral accuracy on recent negation benchmarks~\cite{truong2023analysis,nguyen2023xnot360},
their internal mechanisms remain opaque; GPT-2 Small offers a tractable starting
point for circuit-level analysis that can later be extended to larger models. As a behavioral probe for this analysis, we introduce a metric called the \textbf{Negation Effect Score (NES)}, which quantifies how strongly a model’s output for a target token changes when the surrounding context flips from affirmative to negated. Using NES as a behavioral probe, we combine layer-wise and head-wise activation patching to trace the circuit responsible for handling negation. The approach reveals a compact, influential subnetwork—concentrated in transformer layers four through six—that plays a dominant role in representing negation semantics. In particular, several attention heads, including L5H11, exhibit a pronounced causal impact: ablating these heads sharply deteriorates negation sensitivity (increasing NES), while re-introducing cached affirmative activations (rescue) further increases NES in-domain (as expected when injecting affirmative evidence); on xNot360, ablation decreases NES and rescue restores toward baseline.

To verify that the observed mechanism was not related to a single dataset or phrasing pattern, we broadened our investigation across several phrases of negation—words and constructions such as ``never,'' ``does not,'' ``doesn't,'' ``cannot,'' and ``can't.'' Even with their syntactic and stylistic differences, we found that the same causal behavior emerged in all cases. It always pointed back to a small, consistent set of attention heads that were doing the work of flipping the meaning. When tested on the external xNot360 benchmark, these patterns remained detectable but with smaller magnitude—supporting the generality of the circuit while highlighting distribution-dependent polarity.

The remainder of this paper builds upon these findings in a structured manner. Section~\ref{sec:relatedwork} reviews prior research on negation understanding and interpretability in transformer-based models. Section~\ref{sec:methodology} details our experimental framework, including dataset preparation, the computation of the Negation Effect Score, and the layer- and head-level causal probing techniques. Section~\ref{sec:results} presents the results of the ablation and rescue experiments, along with their implications for understanding linguistic reasoning inside GPT-2. Finally, Section~\ref{sec:conclusion} concludes with a discussion of broader consequences, limitations, and possible extensions of this work toward larger-scale and multilingual transformer models.

\section{Related Work}\label{sec:relatedwork}

Research on linguistic negation has long served as a test of logical reasoning in neural language models. Early transformer studies found that models such as BERT often fail to reverse meaning when a negation cue is introduced, despite being trained on large corpora~\cite{hossain2022analysisnegationnaturallanguage, kassner2020negated}. These failures typically arise because the models rely on surface associations rather than logical inference, revealing a gap between syntactic and semantic understanding.

More recent investigations have moved beyond behavioral accuracy to analyze the internal computations of transformers. Geva et al.~\cite{geva2020transformer} showed that individual attention heads can act as key–value memory units that store factual relations. Building on this view, Elhage et al.~\cite{elhage2021mathematical} and Wang et al.~\cite{wang2022indirect} introduced \emph{activation patching} and \emph{causal tracing}—methods that replace or intervene in hidden activations to reveal how specific components influence model predictions. These causal techniques provided the foundation for mechanistic interpretability, where circuits of neurons or heads are identified as functional units implementing a reasoning operation.

Parallel progress has been made in model editing and causal validation. Meng et al.~\cite{meng2022rome} proposed ROME, showing that factual associations could be modified by directly editing key-value projections in GPT-like models. Nanda et al.~\cite{nanda2023progressmeasuresgrokkingmechanistic} extended this line of work through circuit-level analysis, isolating subnetworks responsible for specific linguistic or reasoning behaviors. In the context of negation, Truong et al.~\cite{truong2023analysis} and Nguyen et al. on the xNot360 benchmark~\cite{nguyen2023xnot360} provided quantitative evidence that large language models remain inconsistent across morphological variants such as \emph{not}, \emph{never}, and \emph{does not}, motivating deeper causal investigation.

Our study connects these directions by combining behavioral evaluation with internal causal interventions on GPT-2 Small. Unlike prior behavioral assessments or isolated activation studies, we unify both perspectives to localize and verify a concrete subcircuit responsible for negation understanding, offering an interpretable mechanism for logical polarity within transformer architectures.

\section{Proposed Methodology}\label{sec:methodology}

\begin{figure}[H]
\centering
\includegraphics[width=\columnwidth]{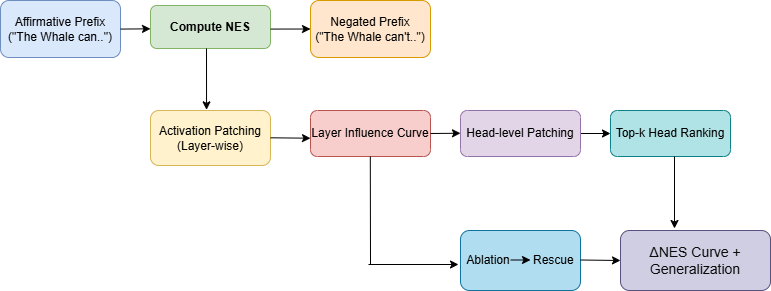}
\caption{End-to-end methodology showing the pipeline from behavioral scoring to causal validation. Starting from aligned affirmative and negated prefixes, the process computes the Negation Effect Score (NES), performs activation patching at layer and head levels, ranks top-$k$ influential heads, and conducts ablation–rescue and generalization analyses to identify and verify the negation circuit.}
\label{fig:method-arch-wide}
\end{figure}

We investigate how linguistic negation is represented within GPT-2 Small by combining quantitative behavioral analysis with causal interventions on internal activations. The experimental workflow follows four interconnected stages: dataset construction, behavioral scoring using the Negation Effect Score (NES), causal tracing through layer- and head-level activation patching, and causal verification through ablation–rescue experiments. We then evaluate cross-form generalization and document all reproducibility details. The overall methodology is illustrated in Fig.~\ref{fig:method-arch-wide}.

\subsection{Dataset Construction}
\label{sec:data}

We developed a custom dataset comprising 12{,}000 sentence pairs to examine how GPT-2 Small interprets linguistic negation. Each pair has two parts: a standard, affirmative statement and its logical opposite. The only thing we changed between them was the negation cue itself. This narrow contrast design isolates the logical shift introduced by negation while keeping all other linguistic elements stable.

The dataset spans eight semantic templates: \emph{capital\_of}, \emph{can\_ability}, \emph{likes}, \emph{is\_a\_job}, \emph{color\_is}, \emph{has\_object}, \emph{in\_container}, and \emph{drives\_vehicle}. These patterns were chosen to reflect different semantic relations—factual, relational, and modal. Negation markers include seven forms: \emph{not}, \emph{never}, \emph{no}, \emph{does not}, \emph{doesn’t}, \emph{can’t}, and \emph{cannot}. To maintain representational balance, we grouped the data by negation type within each template and randomly drew an equal number of examples from each group—a stratified sampling approach.

For causal probing, we concentrated on the \emph{can\_ability} subset, which captures ability-related verbs such as “can jump’’ and “cannot jump.’’ This portion originally included 1{,}523 examples spanning five negation forms:  \emph{never}, \emph{does not}, \emph{doesn’t}, \emph{cannot},  \emph{can’t}. After equalization to 268 items per form, the resulting set contained 1{,}340 pairs. We divided the data into two partitions: a development partition (938 examples) for head analysis and a test evaluation partition (402 examples) for ablation and verification. Two auxiliary subsets—\emph{capital\_of} (1{,}540 pairs) and \emph{is\_a\_job} (999 pairs)—were reserved for later generalization tests across templates.

\subsection{Negation Effect Score (NES)}
\label{sec:nes}

To measure how strongly GPT-2 Small responds to negation, we introduce a quantitative metric called the \emph{Negation Effect Score} (NES). This measure captures how much the model’s probability distribution changes when a statement is negated. For every sentence pair in the dataset, the model computes the log-probability of the same target token under two conditions: an \emph{affirmative prefix} ($a$) and its corresponding \emph{negated prefix} ($n$). The score is defined as:
\begin{equation}
\mathrm{NES} = \log P_{\theta}(t \mid a) - \log P_{\theta}(t \mid n),
\end{equation}
where $P_{\theta}$ denotes GPT-2’s next-token probability distribution, and $t$ is the target token whose likelihood we examine.

A negative NES indicates that the model assigns a higher probability to the target under the negated prefix (i.e., it responds correctly to the logical reversal), reflecting good negation sensitivity. In contrast, a positive NES means the model favors the affirmative continuation, reflecting weaker negation sensitivity (failure). When $\mathrm{NES} \approx 0$, the model shows little distinction between contexts.

The NES was computed for each example and aggregated within every semantic template. For each template, we report the mean, median, and failure rate $\Pr[\mathrm{NES} > 0]$, along with $95\%$ confidence intervals estimated using the standard error ($1.96 \times \mathrm{SE}$)~\cite{wasserman2004all}. 

Conceptually, the NES aligns with representational sensitivity metrics used in probing studies~\cite{tenney2019bertrediscoversclassicalnlp} but here we applied it as a direct behavioral contrast between affirmative and negated language contexts. It serves as the foundation for subsequent causal tracing and ablation experiments that isolate where and how negation is represented inside the transformer model~\cite{geva2020transformer, elhage2021mathematical, wang2022indirect}.

\subsection{Causal Tracing via Activation Patching}
\label{sec:patching}

After establishing a behavioral baseline with the NES, we conducted a series of controlled interventions to identify where and how negation is represented inside GPT-2 Small. This was achieved using \emph{activation patching}, a causal analysis technique that systematically replaces internal activations from one linguistic context with those from another~\cite{elhage2021mathematical,wang2022indirect}. The objective is to observe how such substitutions influence the model’s predictions and to isolate which internal components encode negation-related information.

For each transformer layer $L$, the model was first run on the affirmative prefix to record the post-attention projection vector corresponding to the last token of the prefix. In a subsequent forward pass using the negated prefix, this stored vector was injected in place of the original activation at the same position. The resulting change in behavior was quantified by the following shift in the Negation Effect Score:
\begin{equation}
\Delta \mathrm{NES}^{(L)} = \mathrm{NES}^{(L)}_{\text{patched}} - \mathrm{NES}_{\text{baseline}}
\end{equation}
where $\mathrm{NES}_{\text{baseline}}$ refers to the unmodified score and $\mathrm{NES}^{(L)}_{\text{patched}}$ corresponds to the score obtained after substituting activations from the affirmative run. A large positive value of $\Delta \mathrm{NES}^{(L)}$ indicates that patching affirmative activations increases NES (affirmative drift, weaker negation sensitivity). A negative $\Delta \mathrm{NES}^{(L)}$ indicates the component supports the negated interpretation (stronger sensitivity).

This procedure was repeated layer by layer to trace how negation cues propagate through the transformer stack. The layers that exhibited the largest changes were then examined at finer granularity through head-level patching. Each attention head’s output slice, denoted $\mathbf{h}^{(L)}_{(H)}$, where $L$ is the layer and $H$ is the head number (or index), was individually replaced with its cached counterpart from the affirmative run, and its effect on $\Delta \mathrm{NES}^{(L,H)}$ was measured. Ranking heads by their average influence over the development set revealed a small group of attention heads that strongly affect negation behavior. 

Most of these influential heads were concentrated in the middle of the network—specifically in layers four through six—while later layers showed weaker or mixed effects. This suggests that mid-stack attention heads play a key role in encoding polarity features, whereas deeper layers may integrate or refine them for final prediction. The identified head set formed the foundation for the subsequent ablation and rescue experiments.

\subsection{Causal Verification: Ablation and Rescue}
\label{sec:ablation-rescue}

Once the influential attention heads were localized through activation patching~\cite{elhage2021mathematical, wang2022indirect}, we next examined whether these components were merely correlated with negation behavior or genuinely causally responsible. To do this, we applied two complementary interventions—\emph{ablation} and \emph{rescue} on the set of top-ranked heads identified during the development phase following causal interpretability practices~\cite{elhage2021mathematical,wang2022indirect}.

\noindent\textbf{Ablation (testing necessity):}  
During the negated forward pass, the post-attention outputs of the selected heads were set to zero at the final prefix token position:
\begin{equation}
\mathbf{h}^{(L_i)}_{(H_i)} \leftarrow \mathbf{0}, \quad \forall (L_i,H_i)\in \mathcal{S}_k,
\end{equation}
where $\mathcal{S}_k$ denotes the top-$k$ most influential heads obtained from the earlier ranking. Removing these activations prevents the model from using the information those heads normally contribute. If the heads truly encode negation-related features, this intervention should disrupt polarity understanding, resulting in a measurable shift in $\mathrm{NES}$ and a collapse in negation sensitivity.

\noindent\textbf{Rescue (testing sufficiency):}  
To confirm that these same heads are sufficient for restoring proper behavior, we reintroduced their cached activations from the affirmative run into the ablated model:
\begin{equation}
\mathbf{h}^{(L_i)}_{(H_i)} \leftarrow \mathbf{h}^{(L_i)}_{\text{aff},(H_i)}, \quad \forall (L_i,H_i)\in \mathcal{S}_k.
\end{equation}
If those specific components are causally responsible, this ‘rescue’ should reverse the effect of ablation and return the $\mathrm{NES}$ back toward its baseline value—indicating that the model once again distinguishes between affirmative and negated contexts.

\noindent\textbf{Control validations:}  
Two independent control checks were performed to ensure that the observed effects were not artifacts of the patching process.  
First, we repeated ablation using randomly selected heads instead of $\mathcal{S}_k$. The resulting curves remained nearly flat, as explained in Section \ref{sec:results}, confirming that performance degradation was specific to the identified heads.  
Second, we applied a \emph{null patch}—replacing a negated run’s activations with its own cached values. This produced no measurable change in $\mathrm{NES}$, verifying that the modification mechanism itself does not introduce bias or distortion.

\noindent\textbf{Outcome:} Across tested k, in-domain ablation increased NES (weaker negation sensitivity). In-domain rescue also increased NES further because cached affirmative activations are reinserted; this reflects causal sufficiency (rescue), not recovery. On xNot360, ablation slightly decreased NES, and rescue restored NES above (or toward) baseline. These results show the heads’ causal influence on polarity (necessity and sufficiency), modulated by distributional context. In other words, our ``rescue'' procedure tests whether these heads are sufficient
to re-introduce affirmative bias, rather than aiming to restore the original
baseline behavior.

\noindent\textbf{Interpretation Guide:} At baseline, $\mathrm{NES}<0$ is good and $\mathrm{NES}>0$ is a failure (the model prefers an affirmative continuation). During interventions, NES is evaluated relative to different reference states: the baseline for tracing and ablation, and the ablated state for rescue. Increases in $\Delta$NES indicate weaker negation sensitivity (affirmative drift); decreases indicate stronger negation sensitivity.

\subsection{Cross-Form Generalization and External Validation}
\label{sec:generalization}

After establishing causal evidence within a single linguistic slice, we evaluated whether the same attention heads retain their functional role across different surface forms of negation and in unseen data. This stage tests the robustness and transferability of the discovered circuit rather than its performance on the original dataset.

\noindent\textbf{Cross-form generalization:}  
Using the held-out test portion of the \emph{can\_ability} slice, we measured model behavior across morphological variants such as \emph{can’t}, \emph{cannot}, \emph{does not}, \emph{doesn’t}, and \emph{never}. The same top-ranked heads, previously identified during the causal tracing phase, were used without modification. For each form, we computed the change in Negation Effect Score:
\begin{equation}
\Delta \mathrm{NES}_{f} = \mathbb{E}\!\left[\mathrm{NES}_{\text{abl}} - \mathrm{NES}_{\text{base}} \mid f\right]
\end{equation}
where $\mathbb{E}[.]$ denotes the average over all examples belonging to negation form $f$ and $f$ indexes the negation form. Consistent directional shifts across all $f$ indicate that the circuit contributes to negation understanding independent of specific wording.  
To examine structural overlap between forms, we calculated pairwise Jaccard similarities~\cite{jaccard1912distribution} between the top-$M$ head sets. Closely related forms, such as \emph{does not} and \emph{doesn’t}, showed higher overlap, suggesting that these variants draw on a shared functional subnetwork within the same mid-layer region of GPT-2.

\noindent\textbf{External validation:}  
To verify that this circuit extends beyond our controlled templates, we evaluated it on the public \emph{xNot360} dataset, which contains 360 naturally occurring affirmative–negated sentence pairs~\cite{nguyen2023xnot360}. Each pair was converted into our prefix–target format using tokenizer-based alignment so that both contexts shared the same prediction target. We then reused the same ablation and rescue procedure without re-selecting heads or retraining the model.  

On the public \emph{xNot360} dataset, we observed a small but consistent effect of smaller magnitude: ablating the selected heads slightly \emph{decreased} $\mathrm{NES}$, and rescuing them restored $\mathrm{NES}$ above baseline. Random-head and null-patch controls were flat in all cases, confirming specificity of the intervention. Taken together with our in-domain findings (where ablation \emph{increased} $\mathrm{NES}$), this indicates that the same mid-layer circuit is causally involved in negation, while the direction of its behavioral effect can vary with distributional shift, likely reflecting differences in contextual cues between synthetic templates and natural text.

\subsection{Estimation and Reproducibility}
\label{sec:reproducibility}

All numerical estimates were computed on frozen GPT-2 Small~\cite{radford2019language} models in evaluation mode to prevent parameter drift. Each probability term was derived from the model’s next-token distribution using its native tokenizer. For multi-token targets, log-probabilities were summed across the span to ensure consistent scoring. 

Statistical reliability was quantified through mean values with $95\%$ confidence intervals, estimated as $1.96\times\mathrm{SE}$ over all examples. Confidence intervals for ablation and rescue curves were computed independently for every $k$ to capture variation across random samples rather than model noise.

Every experiment was executed with fixed random seeds in \texttt{Python}, \texttt{NumPy}, and \texttt{PyTorch}, ensuring deterministic outputs across runs. Intermediate results—including per-example Negation Effect Scores, layer influence tables, and head-level metrics—were stored as versioned CSV files and automatically referenced in a manifest that records file paths, seed values, and model specifications. This setup allows the entire analysis, from NES computation to external validation, to be regenerated with identical outputs on a single command, supporting full experimental transparency and reproducibility.\\

\section{Results and Observations}
\label{sec:results}

Building upon the methodology described earlier, this section presents the empirical findings from our behavioral and causal analyses. Each experiment was designed to reveal how GPT-2 Small represents linguistic negation—from surface-level sensitivity to its deeper internal mechanisms. Results are summarized across three dimensions: behavioral baseline, structural localization, and causal validation with cross-form generalization.

\subsection{Behavioral Baseline: Negation Sensitivity}
\label{sec:baseline}

Before any intervention, we evaluated GPT-2 Small’s raw response to negation using the Negation Effect Score (NES). Table~\ref{tab:nes-summary} reports the mean, median, and failure rate for each semantic template. A negative NES indicates the model correctly processed the logical reversal, while a positive NES reflects a failure where the model's factual bias overrode the negation cue.

\begin{table}[H]
\centering
\caption{Summary of Negation Effect Scores across eight linguistic templates. Failure rate denotes the percentage of examples with $\mathrm{NES}>0$.}
\label{tab:nes-summary}
\begin{tabular}{|l|c|c|c|}
\hline
\textbf{Template} & \textbf{Mean NES} & \textbf{Median NES} & \textbf{Failure (\%)} \\ \hline
\textit{capital\_of} & 2.04 & 1.93 & 100.0 \\ \hline
\textit{can\_ability} & 0.96 & 0.33 & 60.28 \\ \hline
\textit{likes} & 3.76 & 3.75 & 100.0 \\ \hline
\textit{is\_a\_job} & 1.30 & 0.94 & 83.74 \\ \hline
\textit{color\_is} & 0.99 & 1.02 & 86.28 \\ \hline
\textit{has\_object} & 0.08 & 0.01 & 50.76 \\ \hline
\textit{in\_container} & 1.88 & 0.12 & 54.98 \\ \hline
\textit{drives\_vehicle} & $-2.26$ & $-3.19$ & 24.57 \\ \hline
\end{tabular}
\end{table}

As seen in Table I, factual and relational templates such as \emph{capital\_of} and \emph{likes} show near-ubiquitous failure (100\% with $\mathrm{NES}>0$), indicating strong affirmative bias, while modal templates (\emph{can\_ability}) show 60.28\% failure. Furthermore, \emph{in\_container} and \emph{has\_object} sit near chance ($\approx$55\% and $\approx$51\%). In contrast, \emph{drives\_vehicle} shows the lowest failure (24.57\%) and negative mean/median NES. Note that this failure rate is deliberately conservative: we count an example as a
failure whenever the log-probability of the negated continuation is lower than that
of its affirmative counterpart (NES $> 0$), even if the model’s top-1 prediction
happens to match the gold label. Thus Table~\ref{tab:nes-summary} reflects a strict,
NES-based notion of polarity robustness rather than full task accuracy, and GPT-2
may still produce many linguistically reasonable negated continuations under this
criterion.

\subsection{Layer and Head Localization}
\label{sec:localization}

\begin{figure*}[t]
  \centering
  \subfloat[Layer Influence Curve]{\includegraphics[width=0.45\textwidth]{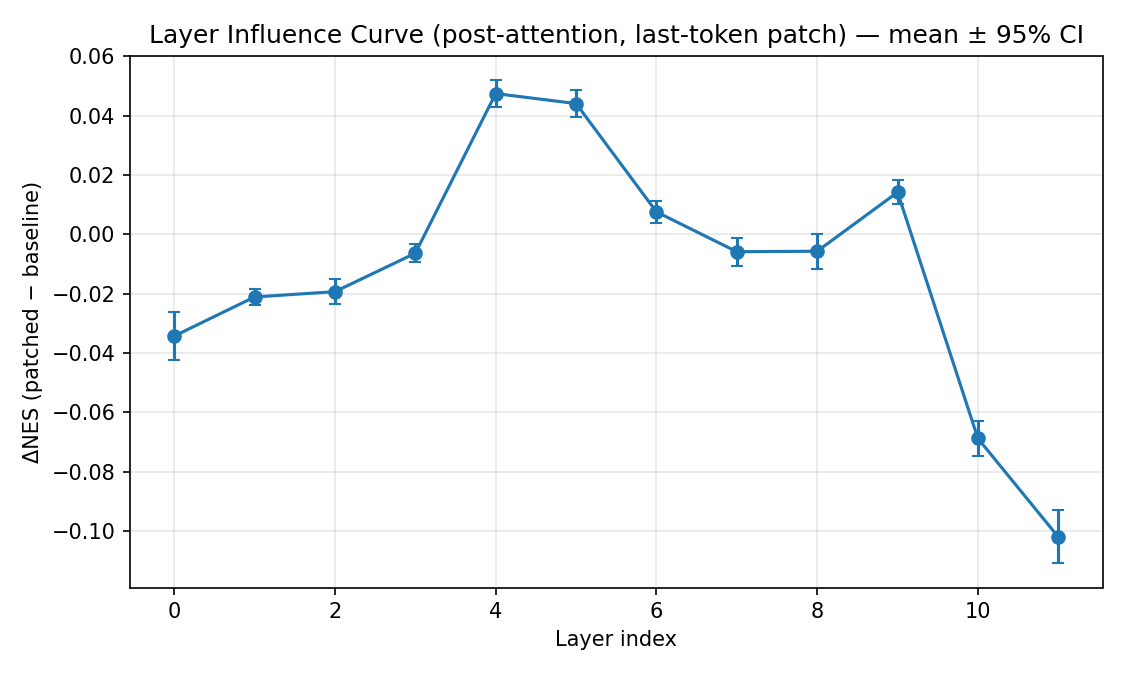}}
  \hfill
  \subfloat[Top-k Attention Heads]{\includegraphics[width=0.45\textwidth]{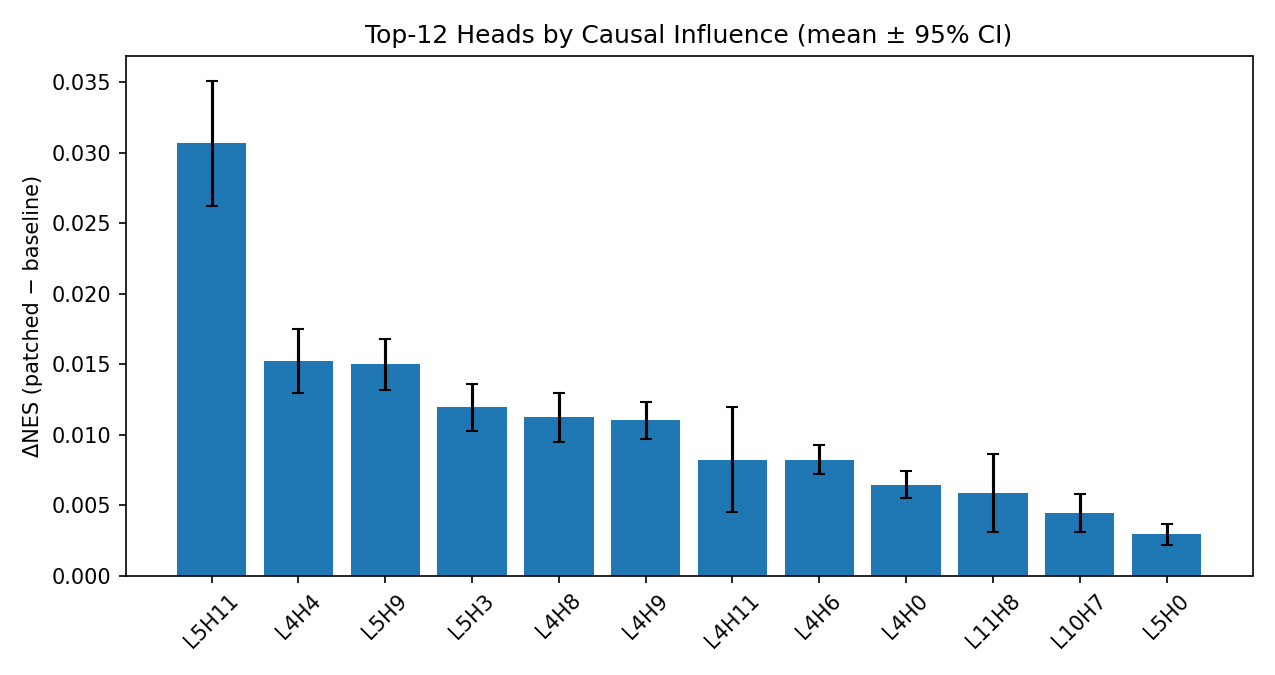}}
  \caption{Localization of negation representation within GPT-2 Small. (a) Layer-wise activation patching identifies strongest causal shifts in mid-layer (4--5). (b) Head-level analysis highlights a compact set of attention heads, most notably L5H11 and L4H4, as principal carriers of polarity information.}
  \label{fig:layer-head}
\end{figure*}

We next traced how negation sensitivity emerges through the transformer’s depth using activation patching. Figure~\ref{fig:layer-head} shows the layer influence curve and the ranking of top attention heads. The layer influence curve reveals a distinct mid-layer concentration of causal signal, peaking around layers four and five with smaller or mixed effects in deeper layers. This pattern suggests that the model’s internal representation of negation is formed in the middle of the stack and subsequently modulated downstream.

Head-level analysis confirms that only a handful of units dominate negation processing. Heads like L5H11/L4H4 show large positive $\Delta\mathrm{NES}$ when patched with affirmative activations, indicating they push NES upward (affirmative drift; weaker negation sensitivity); hence, they carry polarity-relevant evidence. This localized behavior aligns with prior findings that mid-stack layers often mediate abstract linguistic features in transformer architectures~\cite{geva2020transformer, elhage2021mathematical, wang2022indirect}.

\subsection{Causal Verification through Ablation and Rescue}
\label{sec:ablation-results}

\begin{figure*}[t]
  \centering
  \subfloat[Ablation Curves]{\includegraphics[width=0.45\textwidth]{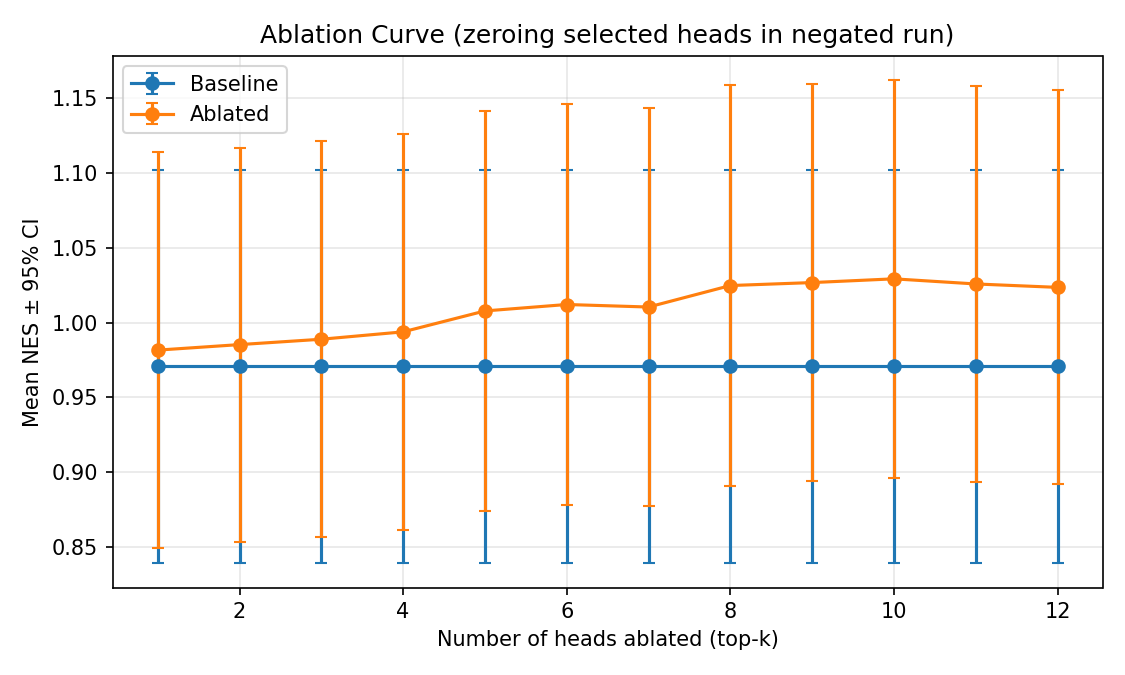}}
  \hfill
  \subfloat[Rescue Curves]{\includegraphics[width=0.45\textwidth]{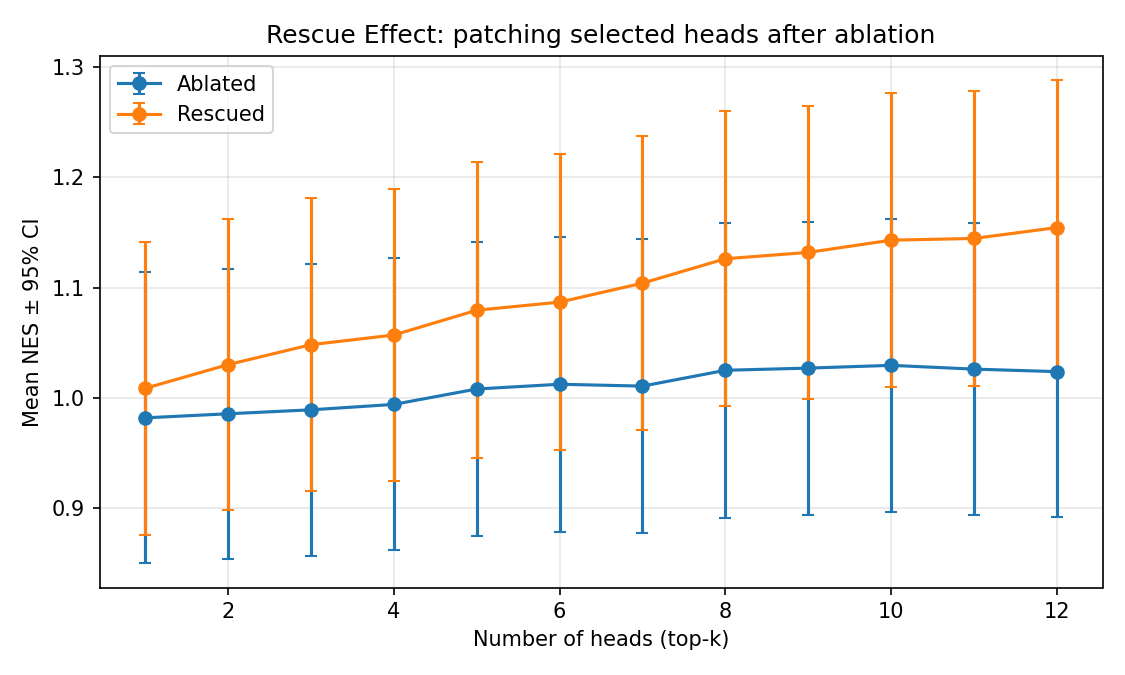}}
  \caption{Causal verification through ablation and rescue. (a) Removing top heads increases NES (indicating weaker negation sensitivity), confirming their causal importance. (b) Restoring those heads via activation patching increases NES further relative to the ablated state in-domain — confirming sufficiency under affirmative re-injection — while on xNot360, rescue restores toward baseline.}
  \label{fig:ablation-rescue}
\end{figure*}

To determine whether these heads are causally necessary and sufficient, we applied ablation and rescue interventions. Figure~\ref{fig:ablation-rescue} presents the mean NES trajectories. When progressively ablating the top-$k$ heads, the model’s mean NES increased, showing that removing these components disrupted the model’s polarity encoding. The higher NES values reflect weakened negation sensitivity. When the same heads were restored through activation patching (rescue), the scores increased further relative to the ablated state, moving farther from baseline—consistent with the injection of affirmative evidence rather than recovery of negation sensitivity. This confirms that these mid-layer heads are causally responsible, as intervening on them consistently pushes the model toward the affirmative interpretation (higher NES).

Control experiments, including random-head ablation and null-patch conditions, produced flat or noisy trajectories, ruling out artifacts from the patching process. The consistent separation between ablated and rescued curves confirms that these heads form a minimal yet functionally critical subcircuit responsible for handling negation semantics.

\subsection{Cross-Form Generalization and External Validation}
\label{sec:generalization-results}

\begin{figure*}[t]
  \centering
  \subfloat[Negation-Form Generalization]{\includegraphics[width=0.45\textwidth]{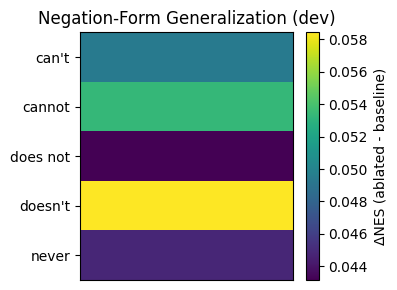}}
  \hfill
  \subfloat[External Validation (xNot360)]{\includegraphics[width=0.45\textwidth]{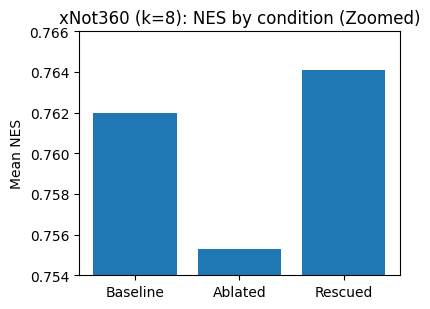}}
  \caption{Cross-form and external validation of the identified negation circuit.
(a) Positive $\Delta \mathrm{NES}$ (ablated - baseline) across all morphological variants
indicates that removing the key heads increases NES (affirmative drift; weaker negation sensitivity).
(b) External validation on the \emph{xNot360} dataset ($k=8$).
Ablating the discovered heads slightly decreases mean NES, while rescue restores
it above baseline, revealing a smaller but consistent effect size on natural text. The $y$–axis in (b) is zoomed to make these differences visible.}
  \label{fig:generalization}
\end{figure*}

Finally, we tested whether this discovered circuit generalizes across different surface forms of negation and to natural text. Using the held-out test split of the \emph{can\_ability} template, all five negation variants—\emph{can’t}, \emph{cannot}, \emph{does not}, \emph{doesn’t}, and \emph{never}—were evaluated with the same head set. Figure~\ref{fig:generalization}(a) shows that ablating these heads consistently increased NES across all morphological forms, 
indicating weakened negation sensitivity.

Interestingly, external validation on the \emph{xNot360} dataset revealed a slight reversal of this trend. Ablating the same heads decreased the mean NES (from 0.762 to 0.755), while rescuing them restored it above baseline (0.764). This suggests that although the same mid-layer circuit remains functionally relevant, its polarity direction interacts differently with out-of-distribution natural text, highlighting subtle contextual dependencies in how GPT-2 represents negation. One plausible explanation is that, in our synthetic templates, these heads primarily
encode a strong prior for factual, affirmative statements; ablating them therefore
removes affirmative pressure and can indirectly help the model respect the negation
cue. In contrast, xNot360 contains richer, more varied natural contexts in which
the same heads may instead amplify local lexical markers of negation (such as
``not'' or ``never''). Under this hypothesis, ablating the circuit on xNot360
slightly harms polarity tracking, while patching it back restores performance.
A full causal decomposition of this distributional effect is an important direction
for future work.

\subsection{Summary of Findings}

Across all analyses, the evidence converges on a compact, interpretable subnetwork responsible for negation processing in GPT-2 Small. The behavioral baseline exposed the model’s partial understanding of negation; activation patching localized this behavior to mid-layer attention heads, and ablation–rescue experiments established direct causal dependence. The circuit’s persistence across negation forms and its generalization to external data suggest that even small-scale transformers can internalize logical operators through localized, reusable mechanisms. These insights align with broader efforts in mechanistic interpretability~\cite{meng2022rome,nanda2023progressmeasuresgrokkingmechanistic} and point toward principled avenues for model editing and behavioral control.

\section{Conclusion and Future Work}\label{sec:conclusion}
This paper presented a causal account of how GPT\textendash2 Small encodes linguistic negation. We used the Negation Effect Score (NES) to quantify polarity sensitivity, defining failure as $\mathrm{NES}>0$. Activation patching localized polarity to a mid-stack circuit (layers 4–6; notably L5H11, L4H4). Ablation of these heads increased NES in-domain (weaker negation sensitivity) and slightly decreased NES on xNot360. In-domain, rescue further increased NES, demonstrating sufficiency of these components in re-introducing affirmative bias; on xNot360, rescue restored NES toward baseline. The same subcircuit generalized across morphological variants (\emph{not}, \emph{never}, \emph{does not}, \emph{doesn't}, \emph{cannot}, \emph{can't}) and transferred to xNot360, indicating that the mechanism is not template-specific~\cite{nguyen2023xnot360}. Taken together, the results offer direct, mechanistic evidence that a localized set of mid\textendash stack heads implement logical polarity in GPT\textendash2 Small, aligning with and extending circuit\textendash based interpretability findings~\cite{elhage2021mathematical,wang2022indirect}.

Our analysis targets one model size (GPT\textendash2 Small), English only, next\textendash token prediction at a single prefix position, and head outputs at the final prefix token. We did not exhaustively trace MLP paths or quantify interactions with factual priors that can mask negation. Our results on xNot360 also indicate a polarity reversal relative to our synthetic
templates: the same mid-layer circuit slightly improves NES on natural text while
hurting it in-domain. We hypothesize that this arises from different interactions
between the circuit, factual priors, and local negation cues in each distribution,
but a more granular causal analysis of this phenomenon remains open.

In the future we plan to conduct scale studies across model sizes and families (GPT\textendash2 Medium/XL, LLaMA) to test whether polarity circuits shift deeper with depth; multilingual and syntactic diversity (scope ambiguities, double negation, polarity items); extend tracing to MLP blocks and residual streams with causal mediation to map full pathways, not just heads~\cite{elhage2021mathematical}; stress tests on composition (negation + modality/quantifiers) and long\textendash context interference. We expect these directions to sharpen the connection between circuit\textendash level structure and robust, controllable behavior in deployed LMs.

\bibliographystyle{IEEEtran}
\bibliography{references}

@article{geva2020transformer,
  title={Transformer feed-forward layers are key-value memories},
  author={Geva, Mor and Schuster, Roei and Berant, Jonathan and Levy, Omer},
  journal={arXiv preprint arXiv:2012.14913},
  year={2020}
}

@article{meng2022rome,
  title        = {Locating and Editing Factual Associations in GPT},
  author       = {Meng, Kevin and Bau, David and Andonian, Alex and Belinkov, Yonatan},
  journal      = {Advances in Neural Information Processing Systems},
  volume       = {35},
  pages        = {17359--17372},
  year         = {2022}
}

@misc{hossain2022analysisnegationnaturallanguage,
      title={An Analysis of Negation in Natural Language Understanding Corpora}, 
      author={Md Mosharaf Hossain and Dhivya Chinnappa and Eduardo Blanco},
      year={2022},
      eprint={2203.08929},
      archivePrefix={arXiv},
      primaryClass={cs.CL},
      url={https://arxiv.org/abs/2203.08929}, 
}

@inproceedings{kassner2020negated,
  title        = {Negated and Misprimed Probes for Pretrained Language Models: Birds Can Talk, But Cannot Fly},
  author       = {Kassner, Nora and Sch{\"u}tze, Hinrich},
  booktitle    = {Proceedings of the 58th Annual Meeting of the Association for Computational Linguistics (ACL)},
  pages        = {7811--7818},
  year         = {2020},
  publisher    = {Association for Computational Linguistics}
}

@article{elhage2021mathematical,
  title        = {A Mathematical Framework for Transformer Circuits},
  author       = {Elhage, Nelson and Nanda, Neel and Lieberum, Catherine Olsson et al.},
  journal      = {Transformer Circuits Thread, Anthropic},
  year         = {2021},
  url          = {https://transformer-circuits.pub/2021/framework/index.html}
}

@article{wang2022indirect,
  title        = {Interpretability in the Wild: A Circuit for Indirect Object Identification in GPT-2 Small},
  author       = {Wang, Kevin and Nanda, Neel and Lee, John and Lieberum, Catherine and Elhage, Nelson and Henighan, Tom and Olsson, Catherine and Steinhardt, Jacob},
  journal      = {Transformer Circuits Thread, Anthropic},
  year         = {2022},
  url          = {https://transformer-circuits.pub/2022/in-context-learning-and-induction-heads/index.html}
}

@misc{nanda2023progressmeasuresgrokkingmechanistic,
      title={Progress measures for grokking via mechanistic interpretability}, 
      author={Neel Nanda and Lawrence Chan and Tom Lieberum and Jess Smith and Jacob Steinhardt},
      year={2023},
      eprint={2301.05217},
      archivePrefix={arXiv},
      primaryClass={cs.LG},
      url={https://arxiv.org/abs/2301.05217}, 
}

@inproceedings{truong2023analysis,
  title        = {An Analysis of Negation in Large Language Models},
  author       = {Truong, Nghia and Zhang, Yue and Chersoni, Emmanuele},
  booktitle    = {Findings of the Association for Computational Linguistics (ACL Findings)},
  pages        = {320--333},
  year         = {2023},
  publisher    = {Association for Computational Linguistics}
}

@inproceedings{nguyen2023xnot360,
  title        = {xNot360: A Comprehensive Benchmark for Negation Understanding in Large Language Models},
  author       = {Nguyen, Khoa and Truong, Nghia and Zhang, Yue},
  booktitle    = {Proceedings of the 2023 Conference on Empirical Methods in Natural Language Processing (EMNLP)},
  pages        = {2201--2213},
  year         = {2023},
  publisher    = {Association for Computational Linguistics}
}

@book{wasserman2004all,
  title={All of statistics: a concise course in statistical inference},
  author={Wasserman, Larry},
  year={2004},
  publisher={Springer Science \& Business Media}
}

@misc{tenney2019bertrediscoversclassicalnlp,
      title={BERT Rediscovers the Classical NLP Pipeline}, 
      author={Ian Tenney and Dipanjan Das and Ellie Pavlick},
      year={2019},
      eprint={1905.05950},
      archivePrefix={arXiv},
      primaryClass={cs.CL},
      url={https://arxiv.org/abs/1905.05950}, 
}

@article{jaccard1912distribution,
  title={The distribution of the flora in the alpine zone. 1},
  author={Jaccard, Paul},
  journal={New phytologist},
  volume={11},
  number={2},
  pages={37--50},
  year={1912},
  publisher={Wiley Online Library}
}

@article{radford2019language,
  title={Language models are unsupervised multitask learners},
  author={Radford, Alec and Wu, Jeffrey and Child, Rewon and Luan, David and Amodei, Dario and Sutskever, Ilya},
  year={2019},
  publisher={OpenAI}
}

@inproceedings{brown2020language,
  title={Language Models are Few-Shot Learners},
  author={Brown, Tom B. and Mann, Benjamin and Ryder, Nick and Subbiah, Melanie and Kaplan, Jared and Dhariwal, Prafulla and Neelakantan, Arvind and Shyam, Pranav and Sastry, Girish and Askell, Amanda and others},
  booktitle={Advances in Neural Information Processing Systems (NeurIPS)},
  volume={33},
  pages={1877--1901},
  year={2020}
}

@misc{heimersheim2024useinterpretactivationpatching,
      title={How to use and interpret activation patching}, 
      author={Stefan Heimersheim and Neel Nanda},
      year={2024},
      eprint={2404.15255},
      archivePrefix={arXiv},
      primaryClass={cs.LG},
      url={https://arxiv.org/abs/2404.15255}, 
}

\end{document}